\documentclass[11pt,a4paper]{article}

\usepackage{times,latexsym}
\usepackage[T1]{fontenc}
\usepackage[table]{xcolor}
\usepackage[acceptedWithA]{tacl2021v1}

\usepackage{graphicx}
\usepackage{booktabs}
\usepackage{array}
\usepackage{ragged2e}

\usepackage{multirow}
\usepackage{amsmath,amssymb}
\usepackage{microtype}
\usepackage{enumitem}

\usepackage{xspace}

\usepackage{amsmath,amsfonts,bm}

\def\eqref#1{equation~\ref{#1}}

\def\1{\bm{1}}

\DeclareMathAlphabet{\mathsfit}{\encodingdefault}{\sfdefault}{m}{sl}
\SetMathAlphabet{\mathsfit}{bold}{\encodingdefault}{\sfdefault}{bx}{n}

\usepackage{hyperref}
\usepackage{url}
\definecolor{PaperBlue}{HTML}{08457D}
\definecolor{TableHeader}{HTML}{EDF2F6}
\hypersetup{
  hidelinks,
  breaklinks=true,
  pdftitle={PAPER2LLM++: Continual Self-Evolution of LLMs from Research Papers},
  pdfauthor={Hongji Pu, Yilun Zhao, Wenpeng Yin}
}
\newcommand{\PaperTableRef}[1]{\hyperref[#1]{Table~\ref*{#1}}}
\newcommand{\PaperFigureRef}[1]{\hyperref[#1]{Figure~\ref*{#1}}}
\newcommand{\PaperAppendixRef}[1]{\hyperref[#1]{Appendix~\ref*{#1}}}
\newcommand{\TableUncertainty}[1]{_{\textcolor{black!55}{\scriptstyle\pm#1}}}

\newcolumntype{P}[1]{>{\RaggedRight\arraybackslash}p{#1}}

\newcommand{\modelname}{\textsc{Paper2LLM++}\xspace}
\newcommand{\ie}{\emph{i.e.,}\xspace}

\definecolor{BlockGray}{RGB}{246,246,246}

\newcommand{\seqblockrow}[1]{%
  \rowcolor{BlockGray}%
  \multicolumn{8}{@{}l}{\rule{0pt}{2.6ex}\color{PaperBlue}\textbf{#1}}%
}

\title{\modelname: Continual Self-Evolution of LLMs from Research Papers}

\author{
Hongji Pu$^{\dagger}$ \qquad
Yilun Zhao$^{*}$ \qquad
Wenpeng Yin$^{\ddagger}$ \\[3pt]
$^{\dagger}$UIUC \qquad
$^{*}$Yale University \qquad
$^{\ddagger}$Penn State \\[2pt]
\texttt{hongjip2@illinois.edu} \qquad
\texttt{yilun.zhao@yale.edu} \qquad
\texttt{wenpeng@psu.edu}
}

\begin{document}

\maketitle

\begingroup
\endgroup

\begin{abstract}
Research on LLMs continually uncovers model limitations, their causes, and potential solutions. Yet these human discoveries remain largely disconnected from model evolution: an LLM does not automatically learn from new research about its own failures. We introduce \textsc{Paper2LLM++}, a framework for \emph{continual self-evolution of LLMs from research papers}. Rather than treating papers merely as knowledge to retrieve, \textsc{Paper2LLM++} uses the growing literature as a stream of evidence and supervision for model improvement. For each incoming paper, it extracts evidence-grounded findings, tests whether the reported limitation persists in the current model, and, when needed, converts the findings into candidate learning signals. A \textit{try--evaluate--commit} procedure integrates an update only when it improves the targeted behavior without substantially forgetting prior improvements or degrading general capabilities. Across a sequential stream of research-discovered LLM failures, we show that models can progressively incorporate new findings while retaining earlier gains. \textsc{Paper2LLM++} thus takes a step toward closing the loop between \emph{human discovery} and \emph{model evolution}, enabling models to continually learn from research about their own limitations and improvements.\footnote{Data \& Code will be released upon publication.}
\end{abstract}

\section{Introduction}
\label{sec:introduction}

A long-standing goal of AI is to build models that \emph{continue improving after deployment}: recognizing what they still need to learn, acquiring new capabilities, and retaining what they have already learned. Existing approaches enable models to generate their own training data, refine outputs, or adapt from new inputs \citep{wang2023selfinstruct,madaan2023selfrefine,zweiger2025seal}, but a fundamental question remains: \textbf{where should the supervision for continual self-evolution come from?} We argue that scientific papers provide a natural and continually growing source. Research on LLMs constantly discovers new failure modes, identifies when and why they occur, and develops methods for evaluating or mitigating them. Such findings therefore provide more than knowledge to retrieve: they offer evidence about \emph{what a model should learn next}. The challenge is to turn this evidence into model-specific, persistent improvements.

We introduce \modelname, a framework for \textit{continual self-evolution of LLMs from research papers}. For each incoming paper, \modelname extracts evidence-grounded findings about a reported limitation and first verifies whether the limitation still exists in the current model. If so, it translates the finding into complementary forms of synthetic supervision (\ie direct, contrastive, robustness, and process supervision) without training on the paper's evaluation benchmark. Because neither every paper nor every generated supervision signal is necessarily useful, \modelname employs a \emph{try--evaluate--commit} procedure that accepts an update only when it improves the targeted behavior while preserving previously acquired improvements and general capabilities. This creates a continual loop from \emph{human discovery} to \emph{model verification, learning, and evolution}.

We evaluate \modelname~on 30 failure-revealing research papers spanning seven failure families, using Llama-3.1-8B-Instruct~\citep{grattafiori2024llama3} and Qwen2.5-7B-Instruct~\citep{yang2024qwen25}. In individual-paper experiments, 24 of 30 papers produce positive held-out gains. Across continual paper streams, \modelname~consistently improves the targeted behaviors while outperforming prior self-evolving and continual learning approaches in preserving previously acquired improvements and general capabilities. Beyond aggregate performance, our analysis reveals three conditions that determine whether research findings become useful learning signals: they must be \textit{teachable}, meaning they can be translated into concrete and verifiable supervision; \textit{needed}, meaning the reported limitation still exists in the current model; and \textit{retained}, meaning the resulting improvement survives subsequent updates. These findings suggest that effective self-evolution requires not simply learning from more research, but selectively determining \emph{what is worth learning, how to learn it, and whether it remains learned}.

Overall, our contributions are threefold:
\begin{itemize}[leftmargin=*,nosep]
    \item We are the first to formulate \textit{paper-driven LLM self-evolution}, treating scientific discoveries about model limitations as learning supervision for continual model improvement.
\item We propose \modelname, closing the loop from discovery to improvement, enabling evidence extraction, failure verification, supervision construction, and continual adaptation.
    \item Across 30 papers, seven failure families, and two LLMs, we demonstrate continual improvement and identify \textbf{teachability, need, and retention} as key conditions governing when scientific findings can drive effective model self-evolution.
\end{itemize}

\begin{figure*}[t]
    \centering
    \includegraphics[width=\textwidth]{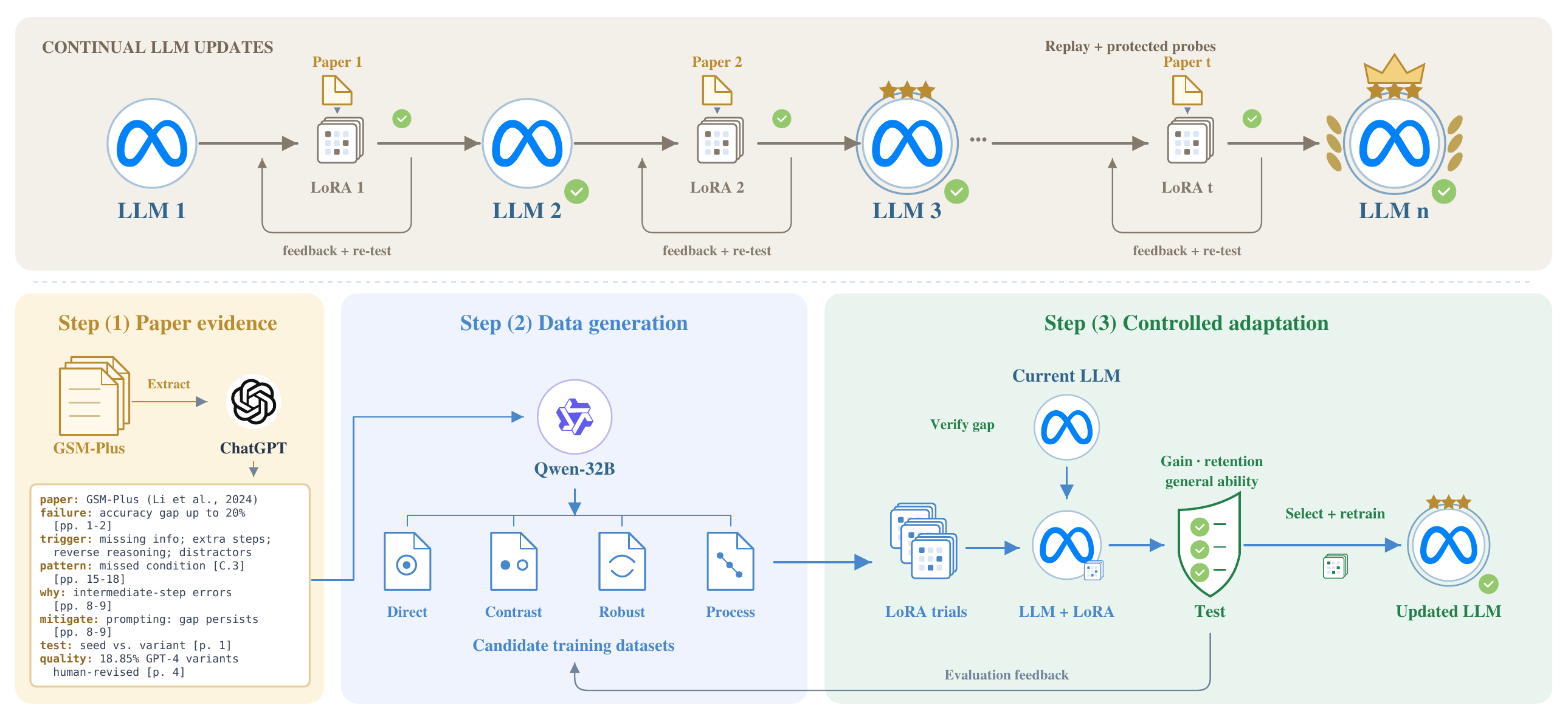}
    \vspace{-8mm}
    \caption{
    \modelname~overview.
    (1) Paper evidence: extract grounded findings and verify the reported gap.
    (2) Data generation: translate confirmed gaps into complementary synthetic supervision.
    (3) Controlled adaptation: select LoRA updates based on target gains, retention, and general capabilities.
    }
    \label{fig:paper2llm_overview}
    \vspace{-2mm}
\end{figure*}

\section{Related Work}
\label{sec:related_work}

\paragraph{Continual and self-evolving LLMs.}
Recent work enables LLMs to continually improve from different forms of incoming experience.
SEAL \citep{zweiger2025seal} learns to convert new passages or task demonstrations into synthetic supervision for parameter updates, but provides limited protection against forgetting.
Recurrent-KIF \citep{feng2025recurrent} and RLEdit \citep{li2025rledit} instead assume that supervision is already available: the former continually learns from labeled task datasets using replay and importance-based parameter fusion, while the latter applies a sequence of explicit knowledge edits through RL-trained parameter updates with memory and locality constraints.
Other approaches evolve the system without modifying the underlying LLM parameters.
G\"odel Agent \citep{yin2025godel} and Darwin G\"odel Machine \citep{zhang2025darwin} use environmental or benchmark feedback to iteratively improve agent code, tools, and workflows, preserving useful solutions through previous implementations or an agent archive.
ACE \citep{zhang2026ace} and LSE \citep{chen2026lse} learn from sequential task experience and feedback by evolving external context, with ACE incrementally maintaining a reusable playbook and LSE retaining and branching previous contexts through an evolution tree.
These approaches therefore primarily study how to continually learn from \emph{given data, tasks, edits, or interaction feedback}, and how to retain the resulting improvements.
In contrast, our \modelname studies a distinct source of continual learning: a stream of \emph{scientific papers}.
This introduces an upstream challenge absent from conventional continual learning: a paper does not directly specify training examples or rewards, but reports findings about model behavior.
\modelname must therefore determine \emph{what} the paper teaches, whether the reported limitation is still present in the current model, translate applicable findings into supervision, and only then incorporate and retain the resulting improvement.

\paragraph{Research papers as external knowledge signals.}
Recent benchmarks treat papers as sources of new knowledge, but primarily specify or evaluate update targets rather than provide an executable procedure for model improvement. ScienceMeter \citep{wang2025sciencemeter} derives scientific claim evaluations from curated papers to measure acquisition, retention, and generalization. DocMEdit \citep{zeng2025docmedit} supplies paired documents containing predefined factual changes, while WikiBigEdit \citep{thede2025wikibigedit} extracts update targets from Wikidata snapshot differences; both leave the actual update procedure to the evaluated method.  Our setting instead asks whether new research papers in stream can be converted into executable supervision, verified against an observed model failure, and incorporated without erasing previously acquired gains.

\section{Problem Formulation}
\label{sec:problem}

We study how an LLM can \emph{continually improve from newly published
research about LLM failures}. Consider a model $M_{t-1}$ that receives a stream of papers
$
    P_t, P_{t+1}, \ldots,
$
where each paper reports an LLM limitation, analyzes when or why
it occurs, and may suggest mitigation strategies.

Our goal is to use each paper as a source of model-maintenance supervision.
A paper, however, is not itself an executable update. The reported failure
may already be absent from $M_{t-1}$; the paper usually does not provide a
training set tailored to $M_{t-1}$; and an update that addresses the new failure
may interfere with previously improved or general capabilities. This gives
rise to three challenges that \modelname addresses:
\vspace{-2mm}
\begin{itemize}[leftmargin=10pt]
    \item \textbf{$\mathcal{C}_1$: Is the reported failure relevant to the current model?}
    A paper reports failures of the models it studied, but $M_{t-1}$ may
    behave differently. We must first assess whether, and to what extent,
the paper-defined failure manifests in $M_{t-1}$.
\vspace{-2mm}
    \item \textbf{$\mathcal{C}_2$: How can a failure-discovering paper be turned into
    executable supervision?}
    If the failure is present, we must identify what it
    teaches, such as the failed behavior, its trigger, and any proposed explanation or
    mitigation, and compile this information
    into synthetic supervision $\widehat{\mathcal{D}}_t$ without training on the evaluation examples.

\vspace{-2mm}
\item \textbf{$\mathcal{C}_3$: How can continual self-evolution preserve previously learned capabilities and general capabilities?}
As papers arrive sequentially, learning from new evidence may overwrite capabilities acquired from earlier papers or degrade the model's general capabilities. Each update must integrate new supervision while preserving prior gains and general model performance.
\end{itemize}

\vspace{-3mm}
\section{Method: \modelname}
\label{sec:method}
\vspace{-2mm}

We describe \modelname (illustrated in \PaperFigureRef{fig:paper2llm_overview}) to address the three challenges: $\mathcal{C}_1$ by extracting supervision and confirming the reported issue; $\mathcal{C}_2$ by generating synthetic supervision data; and $\mathcal{C}_3$ by updating the LLM while maintaining previously learned and general capabilities.

\subsection{Extracting Supervision and Verifying Reported Failures}
\label{sec:confirm_failure}

\paragraph{Extract evidence from the paper.}
Given paper $P_t$, we use gpt-oss-120b~\citep{openai2025gptoss}
\footnote{Temp.~$0$, top-$p=1.0$, 16,384/2,048 input/output tokens.}
to extract a structured record
\[
\begin{aligned}
\mathcal{R}_t = (&\textsc{What},\textsc{Trigger},\textsc{Why},\\
                &\textsc{Mitigation},\textsc{Evaluation}).
\end{aligned}
\]
where  \underline{\textsc{What}}: the behavior or capability reported to fail, such as ``\textit{Models that correctly solve GSM8K
seed questions often fail on minimally perturbed versions of the same problems, producing an accuracy gap of up to 20\%.}'' \cite{gsmplus} (we use this paper as a running example below); \underline{\textsc{Trigger}}: the condition under which the failure appears, such as ``\textit{adding operation}'', ``\textit{reversing operations}'' and ``\textit{distractor insertion}''. \underline{\textsc{Why}}: the paper's own explanation of the failure, when
    explicitly stated. For example ``\textit{often overlook steps or make errors in the middle stages
of the solution}''; \underline{\textsc{Mitigation}}: a proposed solution, such as ``\textit{compositional prompting can help}.'' \underline{\textsc{Evaluation}}: the dataset $D_t$,  and evaluation protocol used
    to assess the failure.

Each field consists of one or more supporting text spans extracted directly from the paper. If a paper does not provide a field (especially \textsc{Why} or
\textsc{Mitigation}), we store \textsc{N/A} rather than infer one. Examples above are
truncated for brevity; a complete evidence-grounded example is provided in \PaperAppendixRef{app:extraction}.

\paragraph{Construct the detection and held-out partitions once.}
The benchmark $D_t$ (further split into $D_t^{\mathrm{dev}}$ and $D_t^{\mathrm{test}}$ by 30/70) associated with  the paper $P_t$ is used for two different purposes:
$D_t^{\mathrm{dev}}$ is the only partition used to decide whether the current LLM
checkpoint exhibits the paper-defined failure, whereas $D_t^{\mathrm{test}}$
is reserved for final assessment.
Thus:
$
    D_t=D_t^{\mathrm{dev}}\cup D_t^{\mathrm{test}}.
$

\paragraph{Verify whether the paper-reported failure exists in the current LLM.}
This step is different from
simply measuring how often  the current LLM $M_{t-1}$ makes mistakes: the goal is to re-assess
the \emph{behavioral effect} reported by the paper using its original evaluation
protocol.
Concretely, we run $M_{t-1}$ on the detection split $D_t^{\mathrm{dev}}$ and
reproduce the comparison used by $P_t$. For papers that report degradation
between two matched conditions, each detection instance consists of a pair
$(x_i, \tilde{x}_i)$, where $x_i$ is the original instance and
$\tilde{x}_i$ is the corresponding instance containing the failure-triggering
change identified by the paper. We measure
\[
\Delta_t =
\operatorname{Score}(M_{t-1}, \{x_i\})
-
\operatorname{Score}(M_{t-1}, \{\tilde{x}_i\}),
\]
using the evaluator specified by $P_t$. A positive $\Delta_t$ therefore means
that the paper-defined trigger degrades the current model's behavior.

For each paper, we follow its original evaluation protocol to reproduce the reported failure; matched pairs are used only when they are part of that protocol. For example,  \citet{gsmplus} pair original GSM8K problems with controlled variations, allowing us to test whether the variations cause a performance drop. We bootstrap the evaluation units to estimate the uncertainty of $\Delta_t$ and obtain a 95\% confidence interval. We confirm the failure only when this interval is entirely above zero, indicating a performance drop under the paper-defined failure condition. Otherwise, we perform \textsc{No Update}; confirmed failures proceed to supervision.

\subsection{Generating Paper-Inspired Learning Data}
\label{sec:synthetic_data}

We do not train the LLM on the benchmark examples in $D_t$. Instead,
Qwen2.5-32B-Instruct~\citep{yang2024qwen25}, as the data generator, receives
only $\mathcal{R}_t$, the structured evidence extracted from the paper, and
generates new examples that instantiate the same behavioral gap. Examples from $D_t$ are never provided to the generator.

The same research finding can be translated into different learning signals,
and it is not clear a priori which form is most effective. We therefore
generate up to four candidate supervision datasets. We illustrate each using
GSM-Plus \citep{gsmplus}, which reports that LLMs can overlook a
newly added condition in a math problem.

\textbullet\enspace\textbf{Direct supervision} teaches the desired behavior directly.
    For example, we generate the problem
    ``\textit{6 boxes contain 12 pencils each, and 3 pencils are removed from each
    box},'' whose correct answer is $54$. This teaches the model to
    incorporate the added condition rather than overlook it.

\textbullet\enspace\textbf{Contrastive supervision} teaches which changes should
    change the answer. For example, we pair
    ``\textit{6 boxes contain 12 pencils each}'' ($72$)
    with
    ``\textit{6 boxes contain 12 pencils each, but 3 pencils are removed from each
    box}'' ($54$).
    The two problems differ only in the added condition, making its effect on
    the answer explicit.

\textbullet\enspace\textbf{Robustness supervision} teaches which variation should
    \emph{not} change the answer. For example, we pair
    ``\textit{3 pencils are removed from each box}''
    with its paraphrase
    ``\textit{3 pencils are taken out of every box}.''
    As the quantities and logic are unchanged, both should retain
    the same answer ($54$).

\textbullet\enspace\textbf{Process supervision} teaches the intermediate reasoning
    implicated by the paper. For the same example, the generated supervision
    explicitly applies the added condition:         ``\emph{
        6$\times$12=72 $\rightarrow$
        72-6$\times$3=54}''.

Not every paper supports all four forms of supervision. For example, process
supervision is generated only when the paper identifies a concrete reasoning
or verification error. We generate all applicable candidates and determine which form improves the current model most.
\begin{table*}[t]
\centering
\footnotesize
\renewcommand{\arraystretch}{1.18}
\setlength{\tabcolsep}{3.8pt}
\begin{tabular}{@{}P{0.19\linewidth}P{0.085\linewidth}P{\dimexpr0.495\linewidth-22.8pt\relax}P{0.23\linewidth}@{}}
\toprule
\rowcolor{TableHeader}
\textbf{Paper Family} & \textbf{\#papers} & \textbf{All Papers / Benchmarks} & \textbf{Typical Failure Pattern} \\
\addlinespace[2pt]
\midrule

Mathematical robustness & 6 &
GSM-Symbolic~\citep{gsm_symbolic},
GSM-Plus~\citep{gsmplus},
MATH-Perturb~\citep{mathperturb},
MathCheck~\citep{mathcheck},
Math-RoB~\citep{mathrob},
MathVista~\citep{mathvista}
& rules, constraints, verification \\
\addlinespace[2pt]

Coding and software & 6 &
HumanEval+~\citep{humanevalplus},
LiveCodeBench~\citep{livecodebench},
CRUXEval~\citep{cruxeval},
EvoEval~\citep{evoeval},
SWE-bench~\citep{swebench},
CrossCodeEval~\citep{crosscodeeval}
& execution, retrieval, interaction \\
\addlinespace[2pt]

Instruction operators & 4 &
Negation~\citep{negation},
Reversal Curse~\citep{reversal},
Instruction Hierarchy~\citep{instructionhierarchy},
Prompt Injection~\citep{promptinjection}
& boundaries and control policies \\
\addlinespace[2pt]

Knowledge and factuality & 4 &
TruthfulQA~\citep{truthfulqa},
SimpleQA~\citep{simpleqa},
GPQA~\citep{gpqa},
MMLU-Pro~\citep{mmlupro}
& retrieval and calibrated reasoning \\
\addlinespace[2pt]

Context grounding & 3 &
Lost-in-the-Middle~\citep{liu2024lost},
RULER~\citep{ruler},
RGB~\citep{rgb}
& retrieval, memory, trust \\
\addlinespace[2pt]

Structured reasoning & 3 &
PlanBench~\citep{planbench},
Self-Critique Plans~\citep{selfcritiqueplans},
GraphQA~\citep{graphqa}
& planning and state verification \\
\addlinespace[2pt]

Agentic reliability & 4 &
CounterBench~\citep{counterbench},
ToRR~\citep{torr},
AgentBench~\citep{agentbench},
ToolBench~\citep{toolbench}
& tools, planning, interaction \\
\addlinespace[2pt]

\bottomrule
\end{tabular}
\vspace{-2mm}
\caption{
Evaluation suite of 30 failure-revealing papers and benchmarks for paper-driven LLM self-evolution.}
\label{tab:dataset}

\end{table*}

\subsection{Continual Self-Evolution}
\label{sec:self-evolving}

To decide which supervision should be used
to evolve the current model, we use a \emph{try--evaluate--commit} procedure:
each candidate is first tested through a small temporary update, and \emph{only a
candidate that improves the new behavior without substantially degrading
previously acquired or general capabilities is allowed to update the model}.

\paragraph{Try each candidate with a temporary update.}
For each candidate supervision dataset $\mathcal{D}_t^{(k)}$, we start from
the same current model $M_{t-1}$ and perform a lightweight LoRA update,
producing a temporary model $M_t^{(k)}$. These updates are used only for
candidate selection and discarded after scoring.

\paragraph{Evaluate candidate supervision and update the model.}
Each temporary model $M_t^{(k)}$ is evaluated along three dimensions:
i) \textbf{New-behavior gain ($G_t^{(k)}$):} improvement on the behavioral gap identified by the current paper;
ii) \textbf{Previous-behavior degradation ($P_t^{(k)}$):} performance loss on behaviors acquired from earlier papers; and
iii) \textbf{General-capability degradation ($Q_t^{(k)}$):} performance loss on general LLM capabilities. We therefore combine these signals as ``utility ($U$)''
\[
    U_t^{(k)}
    =
    G_t^{(k)}
    - \lambda P_t^{(k)}
    - \mu Q_t^{(k)}.
\]
To prevent a candidate with high target gain but substantial regression from
being selected, we additionally require
$
G_t^{(k)} \geq \tau_{\mathrm{gain}}$,
$P_t^{(k)} \leq \tau_{\mathrm{prev}}$, and
$Q_t^{(k)} \leq \tau_{\mathrm{gen}}.
$
All five hyperparameters
$(\lambda,\mu,\tau_{\mathrm{gain}},\tau_{\mathrm{prev}},\tau_{\mathrm{gen}})$
are fixed globally before running the continual-learning experiments and are
shared across all papers and model checkpoints.

Among candidates that pass, we select the one with the highest utility. If none passes, the model is left unchanged. Otherwise, we
discard the temporary update and retrain from
$M_{t-1}$ on the selected supervision using the full budget.

\paragraph{Preserve successful learning for future evolution.}
After committing an update, we retain $M_t$ and update two components.
First, we add examples from the selected supervision $D_t^{(k)}$
to the replay buffer $\mathcal{B}_t$. Future updates mix new
supervision with replay samples to reduce forgetting.
Second, we freeze up to 50 examples from $D_t^{\mathrm{dev}}$
as a protected probe $v_t$.
We sample without replacement using seed 42, stratifying by task
subtype or perturbation category when available and sampling
uniformly otherwise. If fewer than 50 examples are available,
we keep all of them. We then update the probe:
\[
    \mathcal{V}_t = \mathcal{V}_{t-1} \cup \{v_t\}.
\]
These probes are never used for training or synthetic-data
generation. They remain fixed and are re-evaluated during later
updates to detect forgetting of previously learned behaviors.
Without an update, the states remain unchanged:
\[
    M_t = M_{t-1}, \qquad
    \mathcal{B}_t = \mathcal{B}_{t-1}, \qquad
    \mathcal{V}_t = \mathcal{V}_{t-1}.
\]

\section{Experiment Setup}
\label{sec:exp_setup}

We evaluate paper-driven LLM self-evolution under a shared setup that measures
target gains, retention, forgetting, and changes in general capabilities.

\subsection{Evaluation Datasets}

\paragraph{Self-Evolution Paper Stream Datasets.}
We construct an evaluation suite for paper-driven self-evolution from 30
papers that report concrete LLM failures and have corresponding benchmarks
for reproducing them. The resulting suite spans seven families:
\textit{mathematical robustness, coding and software, instruction operators,
knowledge and factuality, context grounding, structured reasoning, and
agentic reliability}. \PaperTableRef{tab:dataset} summarizes each failure family, its high-level
LLM failure pattern, and all source papers and benchmarks.

\paragraph{General-Capability Evaluation Dataset.} General ability is measured on fixed 500-example subsets of HellaSwag~\citep{hellaswag},
WinoGrande~\citep{winogrande}, PIQA~\citep{piqa}, BoolQ~\citep{boolq}, and
ARC-Challenge~\citep{arc}. We report macro-average accuracy across these five datasets
(2,500 instances in total). These reporting subsets are not used
for training or update acceptance.

\begin{figure*}[t]
    \centering
    \includegraphics[width=\linewidth]{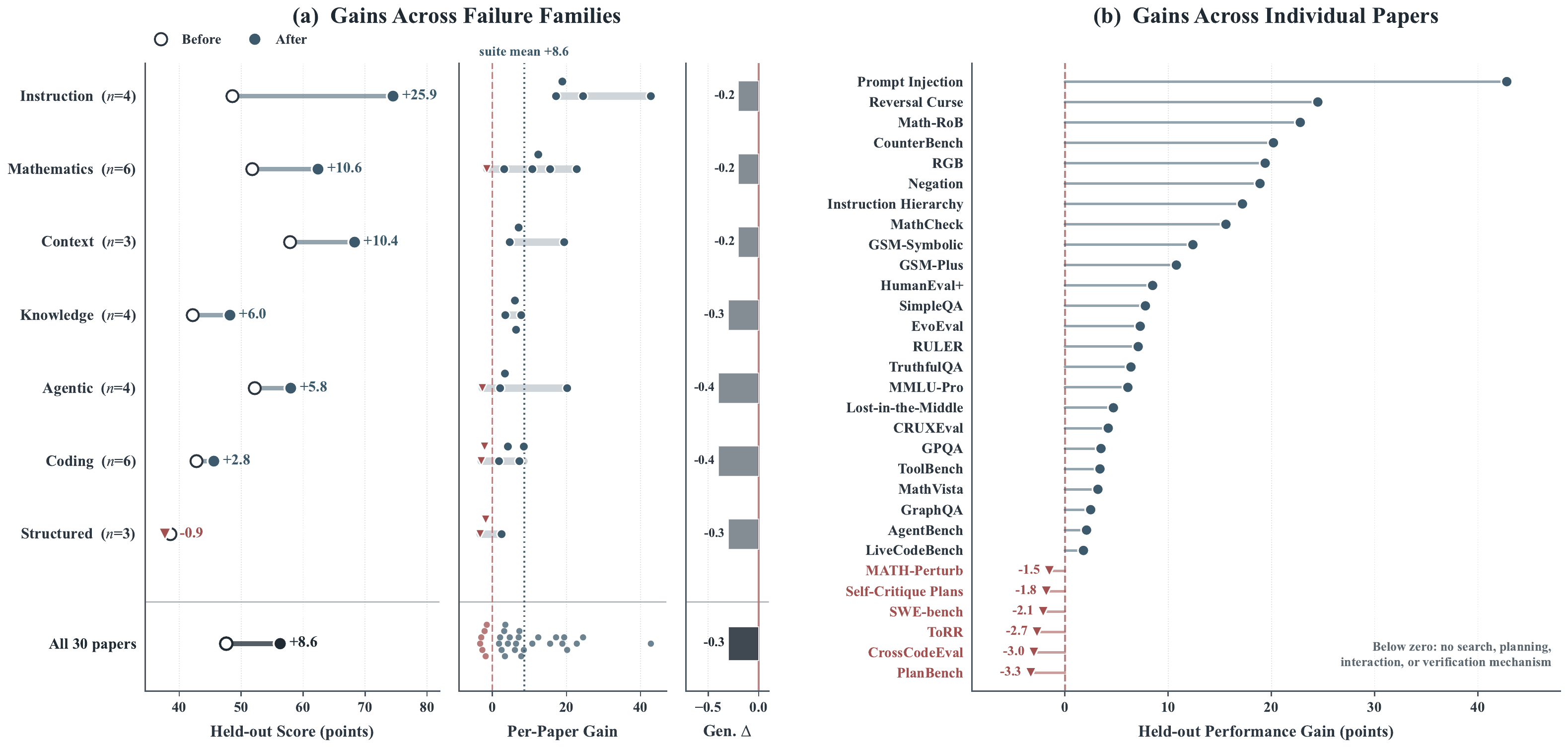}
\vspace{-7mm}
\caption{
Results of Experiment I: individual-paper self-evolution.
(a) Performance gains vary widely even within the same failure family.
(b) 24 of 30 papers show positive held-out gains, while the remaining failures largely involve retrieval, planning, interaction, or verification capabilities beyond the current self-evolution framework.
}
\label{fig:experiment1_patchability}
\vspace{-2mm}
\end{figure*}

\subsection{\modelname \& Baselines}
\modelname~uses Llama-3.1-8B-Instruct
\citep{grattafiori2024llama3} and Qwen2.5-7B-Instruct
\citep{yang2024qwen25}. We use gpt-oss-120b~\citep{openai2025gptoss}
for parsing and knowledge-type classification, and
Qwen2.5-32B-Instruct~\citep{yang2024qwen25} for synthetic data generation.

Regarding baselines, we compare against recent continual-learning and lifelong-editing methods under a unified paper-to-update protocol. All methods process the same papers in the same order and are evaluated with the same held-out procedure; we adapt only how each method consumes the paper-derived signal. Full implementation details and hyperparameters are provided in Appendix~\ref{app:baseline_adaptation}.

\begin{itemize}[leftmargin=*,itemsep=2pt,topsep=3pt]
    \item \textbf{Text-to-LoRA}~\citep{texttolora}:
    we render each extracted paper record as the task description expected by the released hypernetwork, generate one adapter per paper, and merge adapters in paper-arrival order.
    \item \textbf{ELDER}~\citep{elder}:
    we convert each paper's fixed synthetic supervision into a sequence of edits and carry ELDER's mixture-of-LoRA state and router across papers, so that one paper corresponds to one continual editing stage.
    \item \textbf{RLEdit}~\citep{li2025rledit}:
    after pre-training its editor on the authors' original editing data, we apply the frozen editor to the paper-derived edit sets, treating each paper as an editing turn.
    \item \textbf{UltraEdit}~\citep{ultraedit}:
    we replace the editing stream with paper-derived supervision and treat each paper as an editing turn while retaining the released editor and update procedure.
    \item \textbf{StableEdit}~\citep{stableedit}:
    we use UltraEdit's paper-to-edit conversion and carry its running statistics across papers, using general-domain replay only to initialize warm-up statistics.

\end{itemize}

\subsection{Evaluation Setting}
We evaluate self-evolution under two settings:
i) \textbf{individual learning}, where the LLM learns independently from a
single paper; and
ii) \textbf{continual learning}, where the LLM sequentially learns from a
stream of papers. For continual learning, we consider both
\textbf{within-family} streams, containing papers from the same failure family,
and \textbf{cross-family} streams, containing papers across all seven families. The main-table same-family summary reports the coding, structured-reasoning, and agentic-reliability streams. For the continual settings, we report results averaged over three shuffled stream orders and training seeds $\{13,21,42\}$.

We report the following evaluation metrics.
Let $S_i(M)$ be the held-out score for paper $i$, on a 0--100 scale.
Let $M_0$ denote the base model, $M_{i-1}$ the checkpoint before learning
paper $i$, $M_i$ the checkpoint after its update decision, and $M_T$ the
final checkpoint. Define $g_i=S_i(M_i)-S_i(M_{i-1})$.
\begin{itemize}[leftmargin=*,itemsep=2pt,topsep=3pt]
    \item \textbf{Target gain} measures improvement on the current paper's
    benchmark relative to the initial model. For reporting, define
    $G_{\mathrm{base}}=\frac{1}{T}\sum_{i=1}^{T}[S_i(M_i)-S_i(M_0)]$.
    This measures improvement available when each paper is processed,
    including transfer from earlier updates. $g_i=S_i(M_i)-S_i(M_{i-1})$ measures only the current update.
    They coincide for single papers.
    \item \textbf{Retention} measures the fraction of a positive incremental
    gain that survives later updates. For $\mathcal{I}_{+}=\{i:g_i>0\}$,
    \[
    R=\frac{100}{|\mathcal{I}_{+}|}\sum_{i\in\mathcal{I}_{+}}
    \frac{S_i(M_T)-S_i(M_{i-1})}{g_i+\epsilon},
    \]
    where $\epsilon>0$ stabilizes division. Retention is undefined when
    $\mathcal{I}_{+}$ is empty; high retention alone does not imply a large gain.
    \item \textbf{Forgetting} measures the loss between each immediate
    post-update score and its final score:
    $F=\frac{1}{T}\sum_{i=1}^{T}[S_i(M_i)-S_i(M_T)]$.
    Lower is better; negative values indicate improvement.
    \item \textbf{General-capability change} measures collateral effects:
    $\Delta_{\mathrm{gen}}=\frac{1}{5}\sum_{j=1}^{5}
    [A_j(M_T)-A_j(M_0)]$, where $A_j$ is accuracy on general dataset $j$.
\end{itemize}
Gain, forgetting, and general change are reported in score points; retention
is a percentage of acquired gain, not an accuracy.
Means and standard deviations are computed across three orders and seeds.

\section{Experiment and Analysis}
\label{sec:experiments}

We test individual and continual paper learning, then analyze when findings yield lasting gains.

\subsection{Experiment I: Individual-Paper Self-Evolution}
\label{sec:individual}

We first study whether an LLM can evolve by learning from a single research
paper. For each of the 30 papers, we independently start from the same base
LLM, apply \modelname~using only that paper, and evaluate the model
on the held-out benchmark using the same setup and metrics.

\PaperFigureRef{fig:experiment1_patchability}
reports the paper-level results. Among
30 papers, 24 achieve positive held-out gains. The mean across all 30 papers
is +8.6 points, including the six negative outcomes, with a reported mean
general-capability change of -0.3 points. Instruction, mathematics, and context papers improve the most, while coding and
structured reasoning are harder to improve. Overall, we find that learning from a paper does not always lead to a stronger
model: while many papers produce clear gains on their reported behavioral
gaps, others result in little or even negative improvement. The corresponding
family-level values are reported in \PaperAppendixRef{app:complementary_results},
Table~\ref{tab:independent}. This raises an important question:
\textit{What makes a paper effective for self-evolution?} We analyze this question in $\mathcal{Q}_1$.

\begin{table*}[t]
\centering

\scriptsize
\setlength{\tabcolsep}{1.8pt}
\renewcommand{\arraystretch}{1.14}

\begin{tabular*}{\textwidth}{
@{\extracolsep{\fill}}
l
ccc
cccc
@{}
}

\toprule

\multirow{2}{*}{\textbf{Method}}
&
\multicolumn{3}{c}{\textbf{Same-Family Streams}}
&
\multicolumn{4}{c}{\textbf{Mixed 30-Paper Stream}}
\\

\cmidrule(lr){2-4}
\cmidrule(lr){5-8}

&
\textbf{Gain} $\uparrow$
&
\textbf{Ret.} $\uparrow$
&
\textbf{Forget} $\downarrow$
&
\textbf{Gain} $\uparrow$
&
\textbf{Ret.} $\uparrow$
&
\textbf{Forget} $\downarrow$
&
\textbf{Gen. $\Delta$} $\uparrow$
\\

\midrule

\seqblockrow{
Base LLM: Llama-3.1-8B-Instruct~\citep{grattafiori2024llama3}
}
\\

Text-To-LoRA~\citep{texttolora}
&
$+5.8\TableUncertainty{1.0}$
&
$64.9\%\TableUncertainty{5.1}$
&
$4.2\TableUncertainty{0.7}$
&
$+5.1\TableUncertainty{0.9}$
&
$62.6\%\TableUncertainty{4.3}$
&
$5.6\TableUncertainty{0.8}$
&
$-2.0\TableUncertainty{0.5}$
\\

\addlinespace[1pt]

ELDER~\citep{elder}
&
$-0.7\TableUncertainty{0.6}$
&
$79.8\%\TableUncertainty{3.1}$
&
$2.4\TableUncertainty{0.5}$
&
$+3.3\TableUncertainty{0.9}$
&
$72.4\%\TableUncertainty{4.6}$
&
$3.8\TableUncertainty{0.6}$
&
$-1.1\TableUncertainty{0.3}$
\\

RLEdit~\citep{li2025rledit}
&
$\boldsymbol{+7.4}\TableUncertainty{0.9}$
&
$73.5\%\TableUncertainty{4.4}$
&
$3.1\TableUncertainty{0.5}$
&
$\boldsymbol{+7.9}\TableUncertainty{1.1}$
&
$65.2\%\TableUncertainty{5.0}$
&
$4.9\TableUncertainty{0.7}$
&
$-1.6\TableUncertainty{0.4}$
\\

UltraEdit~\citep{ultraedit}
&
$+4.3\TableUncertainty{0.7}$
&
$78.8\%\TableUncertainty{2.9}$
&
$2.7\TableUncertainty{0.6}$
&
$+5.0\TableUncertainty{0.6}$
&
$74.6\%\TableUncertainty{3.2}$
&
$3.5\TableUncertainty{0.5}$
&
$-1.0\TableUncertainty{0.3}$
\\

StableEdit~\citep{stableedit}
&
$+6.3\TableUncertainty{0.5}$
&
$82.6\%\TableUncertainty{3.8}$
&
$2.0\TableUncertainty{0.4}$
&
$+5.7\TableUncertainty{0.8}$
&
$78.1\%\TableUncertainty{4.1}$
&
$3.0\TableUncertainty{0.6}$
&
$-0.8\TableUncertainty{0.2}$
\\

\addlinespace[2pt]

\modelname
&
$+6.8\TableUncertainty{0.7}$
&
$\boldsymbol{86.4\%}\TableUncertainty{2.5}$
&
$\boldsymbol{1.8}\TableUncertainty{0.3}$
&
$+7.1\TableUncertainty{0.7}$
&
$\boldsymbol{83.7\%}\TableUncertainty{3.1}$
&
$\boldsymbol{2.4}\TableUncertainty{0.4}$
&
$\boldsymbol{-0.4}\TableUncertainty{0.2}$
\\

\addlinespace[3pt]

\seqblockrow{
Base LLM: Qwen2.5-7B-Instruct~\citep{yang2024qwen25}
}
\\

Text-To-LoRA~\citep{texttolora}
&
$+4.9\TableUncertainty{0.8}$
&
$72.3\%\TableUncertainty{4.8}$
&
$3.6\TableUncertainty{0.6}$
&
$+5.8\TableUncertainty{0.7}$
&
$58.9\%\TableUncertainty{5.2}$
&
$5.3\TableUncertainty{0.9}$
&
$-1.8\TableUncertainty{0.4}$
\\

\addlinespace[1pt]

ELDER~\citep{elder}
&
$+2.3\TableUncertainty{0.9}$
&
$75.9\%\TableUncertainty{4.0}$
&
$2.8\TableUncertainty{0.4}$
&
$-1.1\TableUncertainty{0.7}$
&
$77.7\%\TableUncertainty{3.3}$
&
$3.2\TableUncertainty{0.5}$
&
$-0.9\TableUncertainty{0.3}$
\\

RLEdit~\citep{li2025rledit}
&
$\boldsymbol{+7.6}\TableUncertainty{1.0}$
&
$76.8\%\TableUncertainty{3.2}$
&
$2.9\TableUncertainty{0.6}$
&
$+6.7\TableUncertainty{1.0}$
&
$71.1\%\TableUncertainty{4.7}$
&
$4.5\TableUncertainty{0.8}$
&
$-1.4\TableUncertainty{0.4}$
\\

UltraEdit~\citep{ultraedit}
&
$+6.0\TableUncertainty{0.8}$
&
$80.9\%\TableUncertainty{3.7}$
&
$2.3\TableUncertainty{0.5}$
&
$+4.4\TableUncertainty{0.8}$
&
$75.8\%\TableUncertainty{2.8}$
&
$3.6\TableUncertainty{0.7}$
&
$-1.2\TableUncertainty{0.3}$
\\

StableEdit~\citep{stableedit}
&
$+5.4\TableUncertainty{0.7}$
&
$83.1\%\TableUncertainty{2.8}$
&
$1.9\TableUncertainty{0.4}$
&
$+6.9\TableUncertainty{0.8}$
&
$80.2\%\TableUncertainty{3.6}$
&
$2.7\TableUncertainty{0.5}$
&
$-0.7\TableUncertainty{0.3}$
\\

\addlinespace[2pt]

\modelname
&
$+7.2\TableUncertainty{0.6}$
&
$\boldsymbol{87.8\%}\TableUncertainty{2.7}$
&
$\boldsymbol{1.5}\TableUncertainty{0.4}$
&
$\boldsymbol{+7.5}\TableUncertainty{0.7}$
&
$\boldsymbol{85.0\%}\TableUncertainty{2.9}$
&
$\boldsymbol{2.1}\TableUncertainty{0.4}$
&
$\boldsymbol{-0.5}\TableUncertainty{0.2}$
\\

\bottomrule

\end{tabular*}

\caption{
Continual self-evolution within (left) and across (right) paper families.
}

\label{tab:sequential_summary}
\end{table*}

\subsection{Experiment II: Continual Self-Evolution Within and Across Families}
\label{sec:continual}

\PaperTableRef{tab:sequential_summary} evaluates continual self-evolution under
both within-family and cross 30-paper stream.

\paragraph{\modelname~achieves a stronger gain--retention balance.}
Across both base LLMs and both stream settings, \modelname~consistently
combines strong acquisition of new behaviors with high retention of previously
learned ones. For example, on the mixed stream, \modelname~achieves gains of
$+7.1$ and $+7.5$ on Llama and Qwen while retaining $83.7\%$ and $85.0\%$ of
acquired gains, respectively. In comparison, RLEdit achieves
gains of similar magnitude ($+7.9$ on Llama and $+6.7$ on Qwen) but substantially lower
retention ($65.2\%$ on Llama and $71.1\%$ on Qwen) and higher forgetting. These
results show that effective self-evolution requires jointly considering
knowledge acquisition and retention rather than optimizing either in
isolation. (Internal ablations: Appendix~\ref{app:complementary_results}, Tables~\ref{tab:internal_coding}--\ref{tab:internal_mixed}.)

\textbf{Longer, mixed-family streams show more forgetting.}
Moving from within-family streams to the mixed 30-paper stream generally makes
continual learning more difficult. For most parameter-updating baselines,
retention decreases and forgetting increases as the stream crosses failure
families. For example, RLEdit's forgetting increases from $3.1$ to $4.9$ on
Llama and from $2.9$ to $4.5$ on Qwen; Text-To-LoRA shows a similar increase
from $4.2$ to $5.6$ and from $3.6$ to $5.3$, respectively. In contrast,
\modelname~shows a smaller increase, from $1.8$ to $2.4$ on Llama and from
$1.5$ to $2.1$ on Qwen, while maintaining strong gains. The mixed streams differ in both family diversity and length, so these
results do not isolate a causal effect of heterogeneity. They show that
\modelname~limits forgetting in the longer, mixed-family setting.

\textbf{\modelname~better preserves general capabilities.}
The mixed-stream setting also evaluates whether continual self-evolution
degrades general LLM capabilities. \modelname~limits the general-capability
change to $-0.4$ on Llama and $-0.5$ on Qwen, compared with substantially
larger degradation for aggressive parameter-updating methods such as RLEdit
($-1.6$ on Llama and $-1.4$ on Qwen) and Text-To-LoRA
($-2.0$ on Llama and $-1.8$ on Qwen). Thus, \modelname's advantage is not simply conservative updating:
it achieves substantial continual gains while limiting both forgetting and
general-capability regression.

\subsection{Analysis: When Does Learning from Research Papers Work?}
\label{sec:analysis}

\begin{figure*}[t]
\centering
\includegraphics[width=0.99\linewidth]{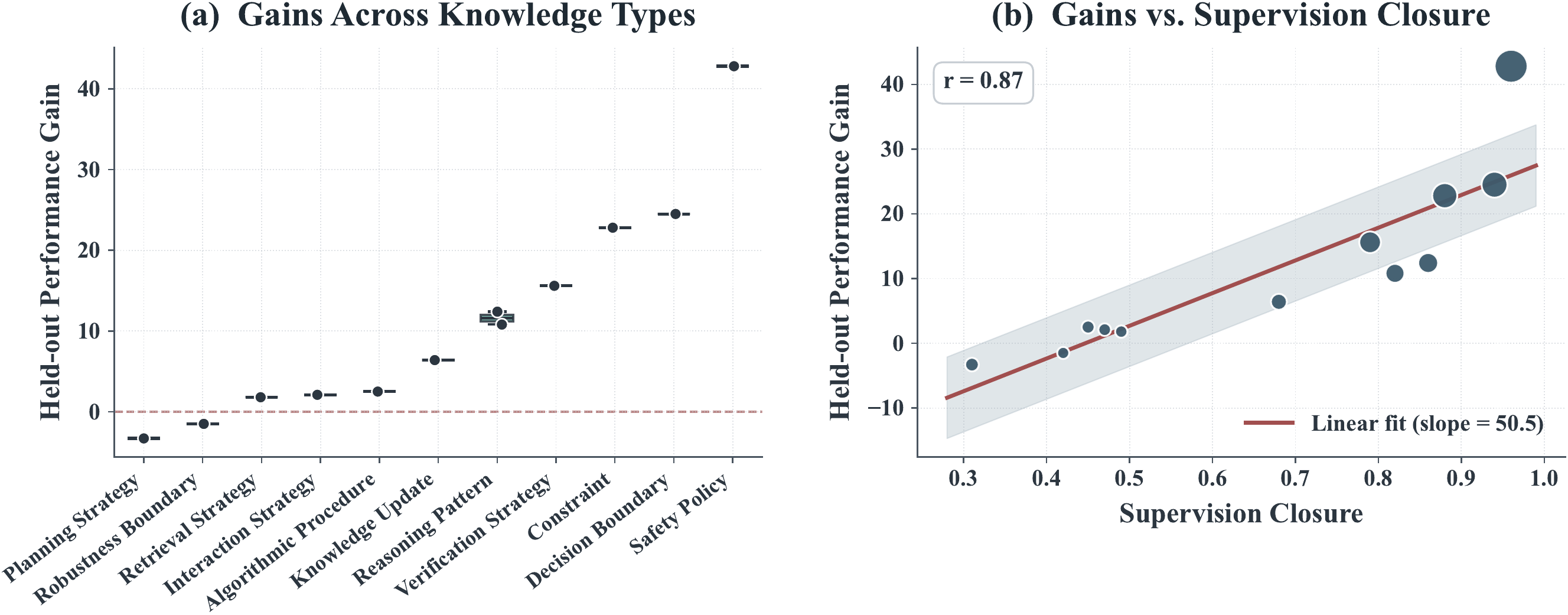}
\vspace{-5mm}
\caption{
Understanding papers' effective signals for LLM self-evolution:
(a) \textbf{Qualitatively:} gains vary across knowledge types.
(b) \textbf{Quantitatively:} supervision closure is associated with held-out gain
($r=0.87$; Appendix~\ref{app:closure}).
}
\label{fig:artifact_analysis}
\vspace{-3mm}
\end{figure*}

The individual-paper and continual experiments establish that LLMs can continually learn from
research papers. We next ask a more fundamental question: \emph{when does a
research finding actually provide a useful signal for self-evolution?}

\begin{figure}
    \centering
    \includegraphics[width=1\linewidth]{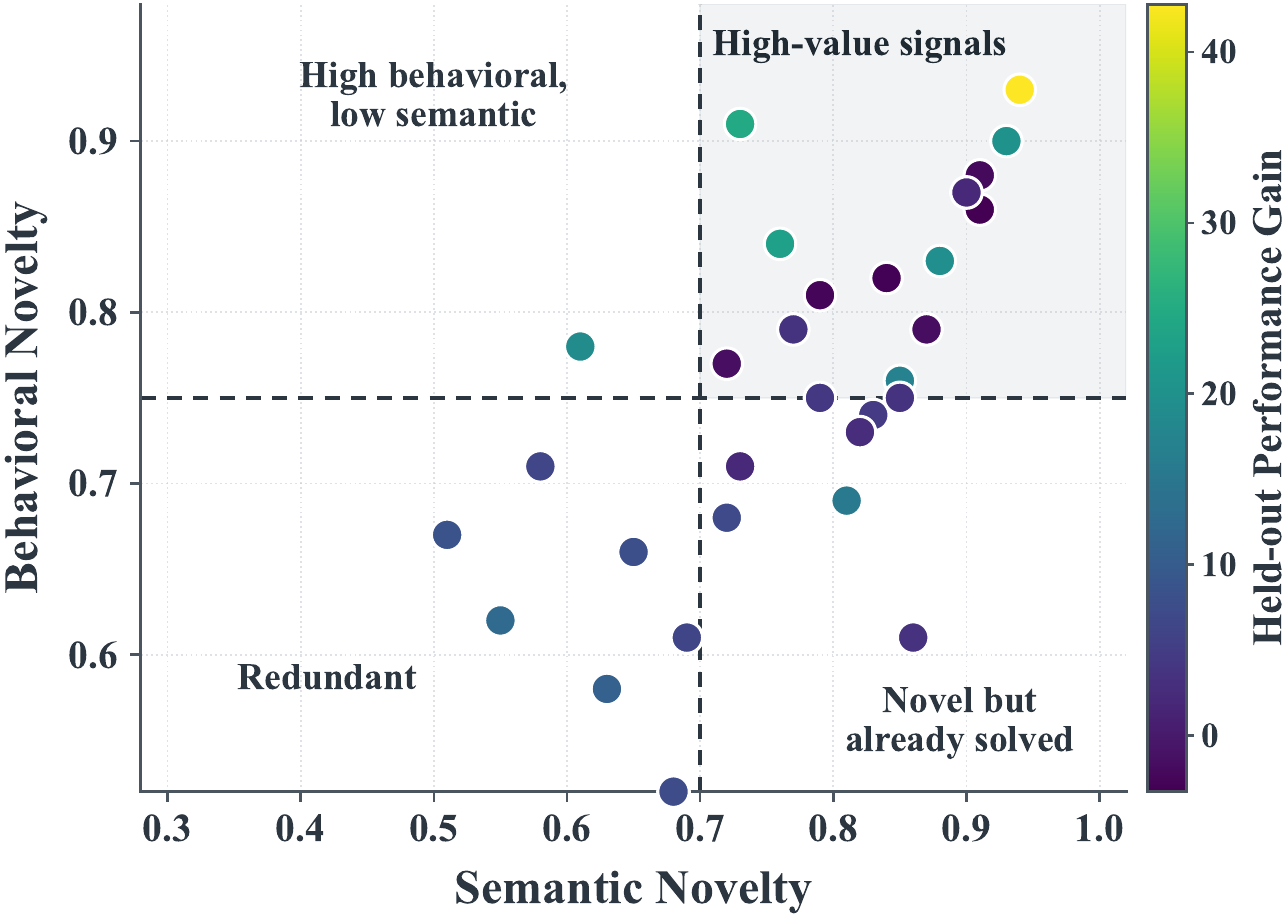}
    \vspace{-8mm}
    \caption{Gain vs. \{semant. novelty, behav. novelty\}}
    \label{fig:novelty}
        \vspace{-3mm}
\end{figure}

\begin{figure}
    \centering
    \includegraphics[width=1\linewidth]{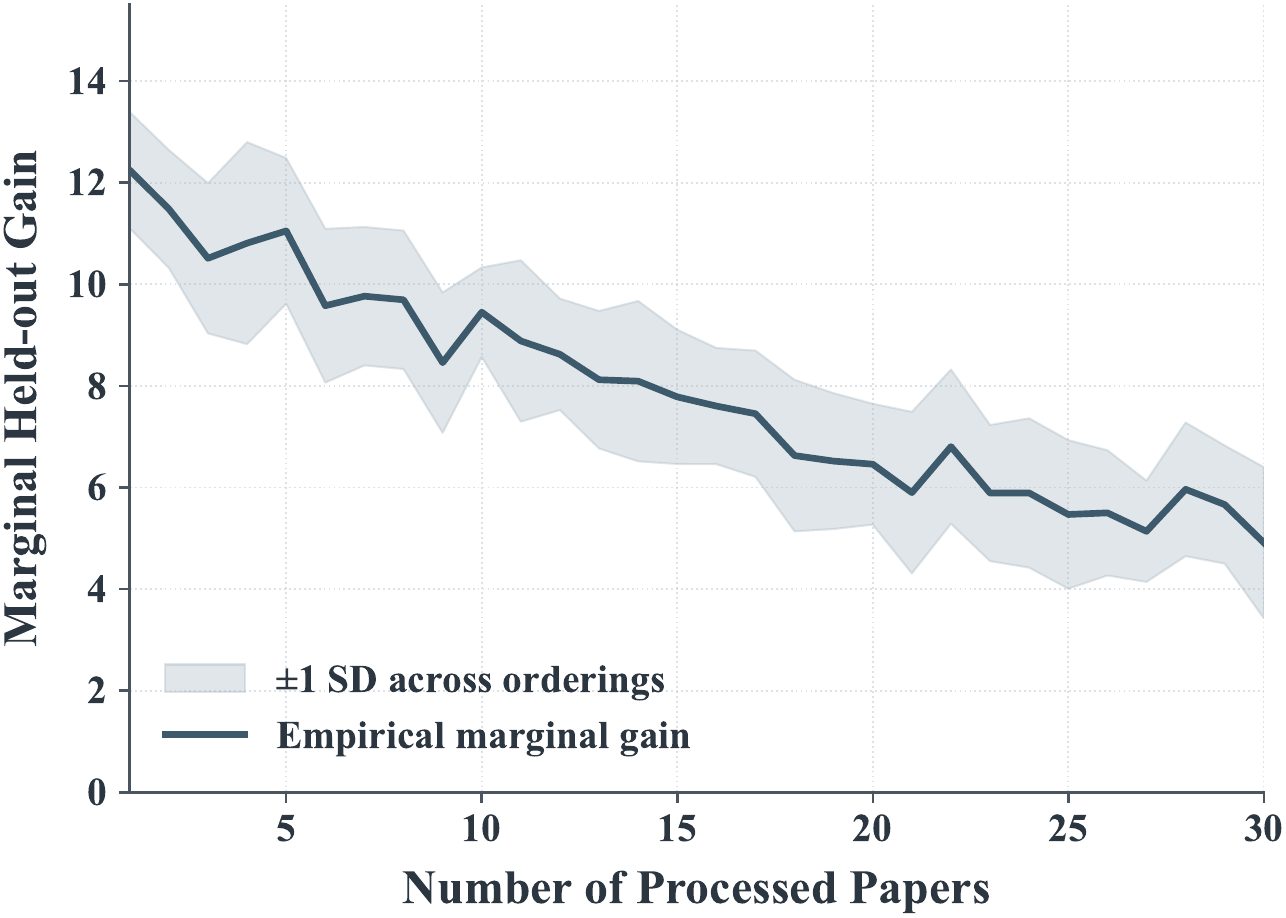}
        \vspace{-8mm}
    \caption{
    Marginal gains decrease but remain positive throughout the 30-paper stream.
    }
    \label{fig:marginal_gain}
        \vspace{-3mm}
\end{figure}

Our analysis suggests that successful paper-driven self-evolution depends on
three conditions. First, the finding must be \textbf{teachable}: it must be
possible to translate the reported insight into concrete supervision that the
model can learn from. Second, it must be \textbf{needed}: the current model
must still exhibit the behavior that the paper aims to improve. Third, what is
learned must be \textbf{retained} as subsequent papers introduce additional
updates. \PaperFigureRef{fig:artifact_analysis}, \PaperFigureRef{fig:novelty}, and
\PaperFigureRef{fig:marginal_gain} examine these conditions across our
30 papers and continual streams. Paper-level values are in
Appendix~\ref{app:complementary_results}, Tables~\ref{tab:artifact_annotations_a}--\ref{tab:artifact_annotations_b}.

\paragraph{$\mathcal{Q}_1$: When Is a Research Finding Teachable?}
We first ask why some papers produce large self-evolution gains while others
produce little improvement. Before inspecting held-out outcomes, we use
gpt-oss-120b~\citep{openai2025gptoss} to annotate each paper by the
\emph{primary knowledge or behavior that must be installed to address its
reported failure}, rather than by its benchmark domain.
The fine-grained labels include reasoning patterns, boundaries, constraints,
verification strategies, retrieval or memory strategies, planning or
interaction strategies, algorithmic procedures, knowledge updates, execution
feedback, and safety policies. For the analysis in
\PaperFigureRef{fig:artifact_analysis}(a), semantically related labels are
grouped into broader knowledge types.

\PaperFigureRef{fig:artifact_analysis}(a) shows that
the answer depends strongly on what the paper asks the model to learn.
Findings involving \emph{constraints, decision boundaries, reasoning patterns, and
verification strategies} generally produce larger gains, whereas findings
requiring \emph{planning, retrieval, or complex interaction} are substantially harder
to acquire.

A common pattern behind these differences is whether the paper's finding can
be turned into concrete training supervision. For example, if a paper
identifies that the model should change its answer under a particular
constraint, we can construct examples that demonstrate when the behavior
should change and automatically check whether the resulting behavior is
correct. In contrast, some findings require capabilities that cannot be
provided by input--output supervision alone, such as searching for unavailable
information, maintaining and updating an external state, interacting with an
environment, or executing and verifying steps.

To describe this distinction, we use each paper's existing \textbf{supervision
closure} annotation, a continuous measure of how readily its finding can be
converted into constructible and verifiable supervision. \PaperAppendixRef{app:closure}
provides a reproducible rubric over five dimensions: explicit target behavior
($c_{\mathrm{target}}$), constructible supervision ($c_{\mathrm{sup}}$), automatic
verification ($c_{\mathrm{ver}}$), local evaluability ($c_{\mathrm{local}}$), and
mechanism sufficiency ($c_{\mathrm{mech}}$). Each dimension is scored continuously
in $[0,1]$: a score of $0$ indicates that the property is absent or cannot be
operationalized from the paper, $0.5$ indicates partial or conditional support,
and $1$ indicates that the property is explicitly and directly satisfied;
intermediate values reflect evidence between these anchors. Closure is then
their unweighted mean:
\begin{equation}
C_t =
\frac{
c_{\mathrm{target}} +
c_{\mathrm{sup}} +
c_{\mathrm{ver}} +
c_{\mathrm{local}} +
c_{\mathrm{mech}}
}{5}.
\end{equation}
The Appendix rubric provides a reproducible interpretation of the existing descriptive scores.

\PaperFigureRef{fig:artifact_analysis}(b) shows a strong relationship between
supervision closure and held-out self-evolution gain ($r=0.87$). Papers whose
findings can be expressed as clear and verifiable learning signals tend to
produce substantially larger improvements. This association is descriptive;
it does not validate the scoring rubric or establish a causal effect.

\emph{Importantly, requiring an additional mechanism does not automatically make a
finding unlearnable. What matters is whether that mechanism can itself provide
a training signal or whether it is the capability that is missing.} For
example, an external checker can make verification useful for learning because
it directly provides correctness feedback. By contrast, when reliable
verification is itself the behavior that the model lacks, showing additional
input--output examples may not be sufficient. Similarly, selecting relevant
evidence already present in the context can be directly supervised, whereas
learning to search for evidence unavailable to the model requires an
additional retrieval process.

\paragraph{$\mathcal{Q}_2$: When Does the Current Model Need the Finding?}
Teachability alone does not make a paper useful for self-evolution. Even a
finding that can be converted into high-quality supervision provides little
value if the current model already handles the reported behavior correctly.
We therefore distinguish what is new in the research literature from what is
still new to the current model.

We characterize \textbf{semantic novelty} by how different a paper-reported
failure is from earlier failure schemas in the stream, and
\textbf{behavioral novelty} by how strongly the current checkpoint still
exhibits the reported failure. Semantic novelty is computed as one minus the
maximum cosine similarity between the current failure-schema embedding and
earlier schema embeddings, while behavioral novelty is the paper-defined
failure severity min--max normalized to $[0,1]$. Full definitions are provided
in \PaperAppendixRef{app:novelty}.

The two forms of novelty capture different situations. A semantically novel
failure may provide little learning value if the current model already handles
it well. Conversely, a familiar failure can remain useful when it exposes an
unresolved weakness of the current checkpoint. The largest gains arise when a
paper provides non-redundant information about behavior the model still needs
to improve. This also motivates a key design choice in \modelname: a new paper does not
automatically trigger an update. \modelname~first verifies whether the reported
behavioral gap remains in the current checkpoint and learns from the paper only
when there is a meaningful gap to address.

\paragraph{$\mathcal{Q}_3$: What Limits Self-Evolution as More Papers Arrive?}
Two factors could limit long-horizon self-evolution. First, the model may
gradually \emph{run out of useful learning signals}: later papers may
increasingly describe behaviors that the model already handles. Second, useful
new findings may continue to arrive, but learning them may overwrite
improvements acquired from earlier papers.

\PaperFigureRef{fig:marginal_gain} provides evidence about the first
possibility. The marginal gain from each additional paper decreases from
approximately 12 points early in the stream to around 5 points near the 30th
paper, but remains positive throughout the studied stream. Thus, although
additional papers provide diminishing returns, useful learning signals do not
appear exhausted at this scale.

The continual-learning results in
\PaperTableRef{tab:sequential_summary} reveal a different challenge: preserving
previously acquired behavior. Parameter-updating methods can achieve gains on
new findings while losing part of earlier improvements, and this effect becomes
more pronounced in cross-family streams. \modelname~substantially reduces this
degradation, maintaining strong gains while achieving higher retention and
lower forgetting than competing parameter-updating approaches across both base
LLMs. This suggests that, at the scale studied here, the central long-horizon
challenge is not simply finding new research signals, but accumulating them
without overwriting prior learning.

\section{Conclusion}
\label{sec:conclusion}
This work introduced \modelname, a framework for continual LLM self-evolution from research papers. Across 30 papers and two LLMs, our results show that research findings can drive persistent model improvements, but not all findings are equally useful. Effective self-evolution requires findings to be \emph{teachable}, \emph{needed}, and \emph{retained}. Our analysis further reveals an important boundary: some failures can be addressed through generated supervision, while others require additional mechanisms such as retrieval, planning, interaction, or verification. Ultimately, scientific literature may serve not only as a record of AI progress, but as a continual source of supervision through which models themselves evolve.

\clearpage

\bibliography{reference}
\bibliographystyle{acl_natbib}

\clearpage
\appendix

\setlength{\textfloatsep}{5pt plus 1pt minus 1pt}
\setlength{\floatsep}{4pt plus 1pt minus 1pt}
\setlength{\intextsep}{4pt plus 1pt minus 1pt}
\setlength{\abovecaptionskip}{3pt}
\setlength{\belowcaptionskip}{0pt}
\newcommand{\AppendixInlineTableCaption}[2]{%
  \refstepcounter{table}\label{#1}%
  \vspace{2pt}\noindent\textbf{Table~\thetable:} #2\par\vspace{5pt}%
}

\section{CATEGORY 1:Replication Details}
\label{app:reproducibility}

\subsection{Model Roles, Paper Input, and Data Isolation}
\label{app:input_protocol}

\paragraph{Model roles.}
Paper parsing and knowledge-type classification use
\textbf{gpt-oss-120b}~\citep{openai2025gptoss}. Synthetic supervision is
generated with \textbf{Qwen2.5-32B-Instruct}~\citep{yang2024qwen25}. The
target backbones are Llama-3.1-8B-Instruct and Qwen2.5-7B-Instruct.
The parser uses temperature $0$, top-$p=1.0$, and 16,384/2,048-token
input/output limits.

\paragraph{Paper input and preprocessing.}
For each paper $P_t$, the parser extracts the paper-reported behavior or
capability failure, its trigger, an explanation and mitigation when explicitly
stated, and the evaluation protocol. The source paper may contain benchmark
descriptions, perturbation rules, failure cases, or representative examples;
these are part of the paper evidence. Source-published examples are removed
from the eligible benchmark pool before the detection/held-out split. Human
preprocessing is restricted to locating benchmark implementations,
standardizing file formats, removing source-published examples from the
eligible benchmark pool, and mapping each paper to its evaluator; humans do
not inspect held-out examples when constructing supervision.

\paragraph{Detection and held-out partitions.}
Each benchmark is divided once using a stratified 30/70 split with seed 42:
\begin{equation}
D_t=D_t^{\mathrm{dev}}\cup D_t^{\mathrm{test}},\qquad
D_t^{\mathrm{dev}}\cap D_t^{\mathrm{test}}=\varnothing.
\end{equation}
Stratification prioritizes benchmark metadata such as perturbation type,
variant, or position bucket when available and otherwise falls back to
evaluator type. The split is fixed across backbones, methods, candidate forms,
stream orders, and training seeds.

\paragraph{Held-out isolation.}
$D_t^{\mathrm{test}}$ is inaccessible during paper parsing, failure
confirmation, synthetic-data generation and filtering, temporary LoRA
training, candidate selection, full training, replay construction, and
stream-order selection. Held-out scores are computed only after the update
decision has been finalized. The generator receives only the extracted paper
record $\mathcal{R}_t$ and never receives raw questions, model responses, or
gold answers from either benchmark partition.

\paragraph{Contamination checks.}
Generated examples are compared with both benchmark partitions using exact
MD5 matching and RapidFuzz string ratio $\geq97$; the checker returns only a
rejection flag and never exposes benchmark content to the generator. Within a
candidate dataset, deduplication uses exact hashes followed by MPNet embedding
cosine similarity $\geq0.93$.

\subsection{Failure Confirmation and Metric Details}
\label{app:detection_metrics}

\paragraph{Failure confirmation.}
The current checkpoint is evaluated on $D_t^{\mathrm{dev}}$ using the
paper-defined comparison. For matched original/perturbed conditions,
\begin{equation}
\Delta_t=\mathrm{Score}(M_{t-1},\{x_i\})-
\mathrm{Score}(M_{t-1},\{\tilde{x}_i\}).
\end{equation}
We bootstrap the paper-defined evaluation units to obtain a 95\% confidence
interval and confirm the paper-defined failure only when this interval lies
entirely above zero. Otherwise, no update is attempted.

\paragraph{Bootstrap implementation.}
We use $B=2000$ bootstrap replicates with random seed 42. The interval is a
95\% percentile interval: after sorting the 2000 bootstrap means, we use the
50th and 1950th order statistics as the lower and upper endpoints. For paired
comparisons, one bootstrap unit is
$100\times(\mathrm{base}-\mathrm{perturbed})$ for one matched pair. For
unpaired accuracy-style evaluations, one unit is
$100\times(1-\mathrm{correct})$ for one example. Resampling is performed over
these paper-defined units.

\paragraph{Retention implementation.}
For paper $i$, let $g_i=S_i(M_i)-S_i(M_{i-1})$. Retention is computed only for
steps with $g_i>0$:
\begin{equation}
R_i=\frac{S_i(M_T)-S_i(M_{i-1})}{g_i+10^{-6}}.
\end{equation}
The ratio is not clipped. Steps with $g_i\leq0$ are excluded from the
retention average, matching the main-text definition.

\subsection{Training, Compute, and Baseline Adaptation}
\label{app:training_details}
\label{app:baseline_adaptation}

\paragraph{Candidate generation.}
For each confirmed paper, \modelname{} constructs the applicable subset of
four candidate forms: direct, contrastive, robustness, and process
supervision. Qwen2.5-32B-Instruct uses temperature $0.7$ and top-$p=0.95$.
The standard generation target is 512 examples and the compute-adjusted target
is 320; generation first produces $1.5\times$ the target count to support
filtering. A candidate is invalidated if fewer than 128 examples survive.
The update budget itself uses 128 accepted new examples together with 64
replay examples, so the larger generation pool does not increase the number
of new examples seen by an update.

\paragraph{Temporary and full updates.}
Each applicable candidate is trained from the same current checkpoint using a
disposable LoRA adapter and the same replay budget. Candidate utility is
\begin{equation}
U_t^{(k)}=G_t^{(k)}-\lambda P_t^{(k)}-\mu Q_t^{(k)},
\end{equation}
subject to the globally fixed acceptance constraints
\begin{equation}
\begin{gathered}
G_t^{(k)}\geq\tau_{\mathrm{gain}},\qquad P_t^{(k)}\leq\tau_{\mathrm{prev}},\\
Q_t^{(k)}\leq\tau_{\mathrm{gen}}.
\end{gathered}
\end{equation}
The selected supervision is retrained from $M_{t-1}$ with the full update
budget.

\begin{table}[!t]
\centering
\scriptsize
\setlength{\tabcolsep}{2.3pt}
\renewcommand{\arraystretch}{0.96}
\begin{tabular}{lcc}
\toprule
Setting & Temporary & Full \\
\midrule
Precision & bfloat16 & bfloat16 \\
Quantization & none & none \\
LoRA rank / alpha & 4 / 8 & 16 / 32 \\
LoRA dropout & 0.05 & 0.05 \\
Learning rate & $10^{-4}$ & $10^{-4}$ \\
Effective batch size & 64 & 64 \\
Optimizer steps & 60 & 1,500 \\
New / replay examples & 128 / 64 & 128 / 64 \\
Sequence length & task dependent & 2,048 / 4,096 / 8,192 \\
Replay capacity & \multicolumn{2}{c}{600 examples} \\
Protected probe & \multicolumn{2}{c}{up to 50 examples} \\
Training / order seeds & \multicolumn{2}{c}{13,21,42 / 101,202,303} \\
\bottomrule
\end{tabular}
\caption{Current training configuration used by the final pipeline. Sequence
lengths are 2,048 for short-form tasks, 4,096 for code and structured tasks,
and 8,192 for long-context tasks.}
\label{tab:fixed_training_config}
\end{table}

The configuration in Table~\ref{tab:fixed_training_config} replaces the
earlier development settings: the final implementation uses 60 temporary
steps, a $10^{-4}$ learning rate for both temporary and full updates, effective
batch size 64, and bfloat16 without quantization.

\paragraph{Replay and protected probes.}
After a successful commit, 64 replay examples are mixed into each future
update. The replay buffer has capacity 600 and preserves a balanced mixture of
60\% prior paper-derived supervision, 20\% matched clean-domain examples, and
20\% general-domain examples. Up to 50 examples from the current
$D_t^{\mathrm{dev}}$ are frozen as a protected probe, sampled without
replacement using seed 42 and stratified by subtype or perturbation category
when available. Protected probes are never used for training or synthetic-data
generation.

\paragraph{General audit and reporting sets.}
Candidate acceptance uses a fixed general-audit set $D^{\mathrm{aud}}$ that is
separate from the final reporting suite. Final general-benchmark results are
macro-averaged over fixed 500-example subsets of HellaSwag, WinoGrande, PIQA,
BoolQ, and ARC-Challenge (2,500 examples total); these reporting subsets are
not used for training or update acceptance.

\paragraph{Continual runs.}
The mixed experiment uses three fixed 30-paper permutations generated with
order seeds 101, 202, and 303, paired with training seeds 13, 21, and 42,
respectively. Every method uses the same three order--seed pairs and no order
is selected or removed after observing results. Same-family experiments use
three fixed permutations for each of the coding, structured-reasoning, and
agentic-reliability streams.

\paragraph{Per-paper optimization accounting.}
Table~\ref{tab:compute_accounting} reports online optimization work performed
per paper by the current implementation. Counts cover optimization during the
paper stream; one-time editor or hypernetwork pre-training is outside this
accounting.

\begin{table}[!t]
\centering
\scriptsize
\setlength{\tabcolsep}{1.8pt}
\renewcommand{\arraystretch}{0.95}
\begin{tabular}{p{0.28\columnwidth}p{0.20\columnwidth}p{0.34\columnwidth}}
\toprule
Method & Online work / paper & Current implementation \\
\midrule
\modelname{} & $\leq4\times60+1500=1740$ steps & Up to four temporary candidates, then one 1,500-step full update; 128 new + 64 replay examples. \\
Seq-LoRA & 1,500 steps & Same 128+64 examples; one full update without probing or rollback. \\
ELDER & $32\times50=1600$ iterations & 32 batches; 50 iterations each; batch 4, rank 8, four LoRA experts, top-2 routing. \\
UltraEdit & 0 gradient steps & Closed-form update over 128 edits. \\
StableEdit & 0 gradient steps & Closed-form update over 128 edits. \\
Text-to-LoRA & 0 gradient steps & One hypernetwork forward from the standardized paper description. \\
RLEdit & not instrumented & 128 edits; current code does not record a directly comparable per-paper count. \\
\bottomrule
\end{tabular}
\caption{Per-paper online optimization accounting. The shared paper stream is
not compute matched: \modelname{} can spend additional optimization on
candidate probing, whereas several editing baselines perform no per-paper
gradient optimization.}
\label{tab:compute_accounting}
\end{table}

\paragraph{Baseline adaptation under the shared paper stream.}
All methods process the same paper order and are evaluated with the same
held-out procedure. Methods that consume example-level supervision receive a
fixed paper-derived set $K_t$ of $N=128$ input--target pairs. Detection and
held-out benchmark examples are never supplied to a baseline. Method-specific
pre-training is completed before the paper stream and is accounted for
separately from per-paper updates.

\begin{table}[!t]
\centering
\scriptsize
\setlength{\tabcolsep}{1.8pt}
\renewcommand{\arraystretch}{0.95}
\begin{tabular}{p{0.23\columnwidth}p{0.69\columnwidth}}
\toprule
Method & Adaptation to the paper stream \\
\midrule
Text-to-LoRA & Render $\mathcal{R}_t$ as the task description; generate one adapter per paper and merge in arrival order. The Qwen2.5-7B hypernetwork is trained on disjoint task descriptions/adapters and frozen before evaluation. \\
ELDER & Convert $K_t$ into edits; one paper is one continual stage; carry mixture-of-LoRA state, router, guided loss, and deferral across papers. \\
RLEdit & Pre-train and freeze the editor on the authors' original editing data, then apply it sequentially to paper-derived edit sets. \\
UltraEdit & Replace the edit stream with $K_t$; retain the released editor, sequential loop, layer choices, and step size. \\
StableEdit & Use the same paper-to-edit conversion as UltraEdit and preserve lifelong-normalization statistics across papers. \\
Seq-LoRA & Use the same LoRA configuration, 128 new examples, 64 replay examples, and 1,500-step full-update budget as \modelname{}, but update after every paper without failure detection, candidate probing/selection, or commit/rollback. \\
\bottomrule
\end{tabular}
\caption{Adaptation of continual-learning and lifelong-editing baselines to
the common paper stream. The interface standardizes paper-derived input
exposure while preserving each method's native update mechanism; it does not
equalize optimization compute.}
\label{tab:baseline_adaptation}
\end{table}

\subsection{Analysis Annotation Protocols}
\label{app:analysis_annotations}

The analysis-only annotations are created before inspecting held-out outcomes
and are never supplied to \modelname{} as update features.

\paragraph{Knowledge type.}
Before inspecting held-out outcomes, gpt-oss-120b assigns each paper the
primary knowledge or behavior that must be installed to address the reported
failure, rather than the benchmark domain. Fine-grained labels include
reasoning patterns, boundaries, constraints, verification strategies,
retrieval or memory strategies, planning or interaction strategies,
algorithmic procedures, knowledge updates, execution feedback, and safety
policies.

\paragraph{Supervision closure.}
\label{app:closure}
Closure is a continuous descriptive annotation of whether a reported failure
can be converted into executable and verifiable supervision. Five components
are scored continuously in $[0,1]$: explicit target behavior, constructible
supervision, automatic verification, local evaluability, and mechanism
sufficiency. A score of $0$ means the property is absent or cannot be
operationalized from the paper, $0.5$ indicates partial or conditional
support, and $1$ means it is explicitly and directly satisfied; intermediate
values reflect evidence between these anchors. The aggregate is
\begin{equation}
C_t=\frac{c_{\mathrm{target}}+c_{\mathrm{sup}}+c_{\mathrm{ver}}+
 c_{\mathrm{local}}+c_{\mathrm{mech}}}{5}.
\end{equation}
These annotations are not supplied to \modelname{} as update features. For
Figure~\ref{fig:artifact_analysis}(b), semantically related fine-grained
labels are merged into the broader knowledge types used in panel (a), and the
reported Pearson $r=0.87$ is computed from the grouped, unrounded analysis
values. The Category~2 table reports raw paper-level annotations and is not the
direct unit of this correlation.

\paragraph{Semantic and behavioral novelty.}
\label{app:novelty}
With $\phi_t=\mathrm{Embed}(\mathrm{FailureSchema}(P_t))$, semantic novelty is
\begin{equation}
N_t^{\mathrm{sem}}=1-\max_{i<t}\cos(\phi_t,\phi_i),\qquad
N_1^{\mathrm{sem}}=1.
\end{equation}
If $\delta_t$ denotes the model-specific paper-defined failure severity,
behavioral novelty is
\begin{equation}
N_t^{\mathrm{beh}}=\frac{\delta_t-\delta_{\min}}{\delta_{\max}-\delta_{\min}}.
\end{equation}
Both quantities are used only for descriptive analysis and are not inputs to
the update procedure.

\subsection{Evidence-Grounded Extraction Example}
\label{app:extraction}

\begingroup
\footnotesize
\setlist[description]{leftmargin=1.7cm,labelwidth=1.55cm,labelsep=0.15cm,itemsep=2pt,topsep=2pt,parsep=0pt}
\begin{description}
\item[\textbf{What}] \textit{Summary:} Models that correctly solve GSM8K seed questions can fail on minimally perturbed versions, with a gap up to 20\%. \textit{Evidence:} ``Our benchmark revealed a gap of up to 20\% between the accuracy reported by the current model and the accuracy observed in our setting, while human performance remains unaffected due to the unchanged inherent difficulty level of the questions.'' (pp.~1--2)
\item[\textbf{Trigger}] \textit{Summary:} Failures are triggered by missing information, additional operations, reversed reasoning, and distractors; adding an operation causes a drop above 40\% for most models except GPT-4 and GPT-3.5-Turbo. \textit{Evidence:} ``Except GPT-4 and GPT-3.5-Turbo, most models, regardless of size and output format, experience an accuracy drop exceeding 40\%.'' (pp.~6--7)
\item[\textbf{Failure pattern}] \textit{Summary:} Errors include operator misuse, omitted conditions, and incorrect use of irrelevant information. \textit{Evidence:} ``Example C.3 shows an example of failure for GPT-3.5-Turbo, where it fails to consider one condition ...''; the condition ``should be considered after step [1] and before step [2].'' (pp.~15--18)
\item[\textbf{Why}] \textit{Summary:} Models can lose intermediate constraints or make mistakes in multi-step reasoning. \textit{Evidence:} ``We find that LLMs often overlook steps or make errors in the middle stages of the solution.'' (pp.~8--9)
\item[\textbf{Mitigation}] \textit{Summary:} Chain-of-thought, least-to-most, program-of-thought, and compositional prompting help but do not eliminate the gap. \textit{Evidence:} ``Although compositional prompting improves performance on seed questions and their variations, it does not bridge the performance gap ...'' (pp.~8--9)
\item[\textbf{Evaluation}] \textit{Summary:} The benchmark evaluates 25 LLMs under four prompting methods and compares solved GSM8K seeds with GSM-Plus variations. \textit{Evidence:} ``Our experiments on 25 LLMs and 4 prompting techniques show that ... their performances are far from robust.'' (p.~1)
\item[\textbf{Data quality}] \textit{Summary:} Human validation remains necessary; annotators revised 18.85\% of GPT-4-generated variants. \textit{Evidence:} ``Overall, human annotators revised 18.85\% of the variations produced by GPT-4, highlighting the importance of human revision.'' (p.~4)
\end{description}
\noindent This example shows that each summarized field is tied to a concrete source span rather than inferred from held-out benchmark outcomes.
\endgroup

\section{CATEGORY 2:Complementary Results}
\label{app:complementary_results}

Unless otherwise noted, model-dependent values in this section are averaged
over Llama-3.1-8B-Instruct and Qwen2.5-7B-Instruct.

\begingroup
\tiny
\setlength{\tabcolsep}{1.6pt}
\renewcommand{\arraystretch}{0.92}
\noindent\resizebox{\columnwidth}{!}{%
\begin{tabular}{llrrrrlrr}
\toprule
Paper & Type & Sem. & Beh. & Cl. & Mech. & Fam. & Gain & Ret. \\
\midrule
GSM-Symbolic & Reasoning & .55 & .62 & .86 & -- & Math & 12.4 & 91\% \\
GSM-Plus & Reasoning & .63 & .58 & .82 & -- & Math & 10.8 & 89\% \\
MATH-Perturb & Robustness & .72 & .77 & .42 & V & Math & -1.5 & 72\% \\
MathCheck & Verification & .81 & .69 & .79 & V & Math & 15.6 & 93\% \\
Math-RoB & Constraint & .76 & .84 & .88 & -- & Math & 22.8 & 94\% \\
MathVista & Multimodal & .86 & .61 & .52 & SV & Math & 3.2 & 80\% \\
HumanEval+ & Verification & .51 & .67 & .78 & V & Code & 8.5 & 88\% \\
LiveCodeBench & Algorithmic & .73 & .71 & .49 & SV & Code & 1.8 & 74\% \\
CRUXEval & Execution & .79 & .75 & .61 & VI & Code & 4.2 & 78\% \\
EvoEval & Reasoning & .68 & .52 & .73 & -- & Code & 7.3 & 86\% \\
SWE-bench & Interaction & .91 & .88 & .35 & SPVI & Code & -2.1 & 68\% \\
CrossCodeEval & Retrieval & .84 & .82 & .38 & SI & Code & -3.0 & 70\% \\
Negation & Boundary & .61 & .78 & .93 & -- & Instr. & 18.9 & 95\% \\
Reversal Curse & Boundary & .73 & .91 & .94 & -- & Instr. & 24.5 & 96\% \\
Instruction Hier. & Constraint & .85 & .76 & .87 & V & Instr. & 17.2 & 92\% \\
\bottomrule
\end{tabular}}
\AppendixInlineTableCaption{tab:artifact_annotations_a}{Paper-level analysis values (1/2). Together with Table~\ref{tab:artifact_annotations_b}, this exposes all 30 raw annotations used by the analysis. Gain is the individual-paper held-out gain; Ret. is paper-level continual retention. Cl.=closure; S/P/V/I denote search/planner/verifier/interaction.}

\noindent\resizebox{\columnwidth}{!}{%
\begin{tabular}{llrrrrlrr}
\toprule
Paper & Type & Sem. & Beh. & Cl. & Mech. & Fam. & Gain & Ret. \\
\midrule
Prompt Injection & Safety & .94 & .93 & .96 & VI & Instr. & 42.8 & 97\% \\
TruthfulQA & Knowledge & .58 & .71 & .68 & S & Know. & 6.4 & 83\% \\
SimpleQA & Knowledge & .65 & .66 & .71 & S & Know. & 7.8 & 85\% \\
GPQA & Reasoning & .77 & .79 & .54 & SV & Know. & 3.5 & 79\% \\
MMLU-Pro & Boundary & .69 & .61 & .62 & -- & Know. & 6.1 & 84\% \\
Lost-in-the-Middle & Retrieval & .83 & .74 & .56 & S & Context & 4.7 & 78\% \\
RULER & Memory & .72 & .68 & .63 & SV & Context & 7.1 & 82\% \\
RGB & Trust boundary & .88 & .83 & .81 & SVI & Context & 19.4 & 91\% \\
PlanBench & Planning & .91 & .86 & .31 & SPVI & Struct. & -3.3 & 65\% \\
Self-Critique Plans & Verification & .87 & .79 & .36 & PVI & Struct. & -1.8 & 70\% \\
GraphQA & Algorithmic & .82 & .73 & .45 & PV & Struct. & 2.5 & 74\% \\
CounterBench & Boundary & .93 & .90 & .91 & VI & Agentic & 20.2 & 95\% \\
ToRR & Reasoning bound. & .79 & .81 & .39 & SPVI & Agentic & -2.7 & 69\% \\
AgentBench & Interaction & .90 & .87 & .47 & SPI & Agentic & 2.1 & 73\% \\
ToolBench & Tool-use & .85 & .75 & .52 & SPVI & Agentic & 3.4 & 76\% \\
\bottomrule
\end{tabular}}
\AppendixInlineTableCaption{tab:artifact_annotations_b}{Paper-level analysis values (2/2). Values are rounded for compact presentation; the reported correlations are computed from the released unrounded analysis data.}

The internal selection-loop ablations in Tables~\ref{tab:internal_coding}--\ref{tab:internal_mixed}
average results over both target backbones. In these tables, Eval is the mean
held-out target score immediately after each paper's update decision; Clean $\Delta$
and Gen. $\Delta$ are final changes relative to the base model on the matched
clean-domain and five-benchmark reporting suites, respectively.

\noindent\resizebox{\columnwidth}{!}{%
\begin{tabular}{lrrrrrr}
\toprule
Coding & Eval & Gain & Ret. & Forget & Clean $\Delta$ & Gen. $\Delta$ \\
\midrule
Base & 42.8 & -- & -- & -- & 0.0 & 0.0 \\
Prompt-only & $43.9\pm.2$ & $+1.1\pm.2$ & $91.8\pm1.6\%$ & $.5\pm.1$ & $-.1\pm.1$ & $-.1\pm.1$ \\
DataSynth-only & $44.5\pm.3$ & $+1.7\pm.3$ & $88.6\pm2.1\%$ & $.8\pm.2$ & $-.2\pm.1$ & $-.2\pm.1$ \\
LoRA-only & $45.6\pm.5$ & $+2.8\pm.5$ & $65.4\pm3.8\%$ & $3.6\pm.4$ & $-1.0\pm.2$ & $-1.2\pm.3$ \\
Generic+LoRA & $45.8\pm.4$ & $+3.0\pm.4$ & $68.9\pm3.2\%$ & $3.2\pm.3$ & $-1.0\pm.2$ & $-1.1\pm.2$ \\
Raw-Gain & $47.4\pm.6$ & $+4.6\pm.6$ & $66.8\pm4.1\%$ & $3.8\pm.5$ & $-1.2\pm.2$ & $-1.4\pm.3$ \\
CBA-selection & $46.7\pm.4$ & $+3.9\pm.4$ & $93.6\pm1.8\%$ & $1.9\pm.2$ & $-.8\pm.1$ & $-.6\pm.2$ \\
\bottomrule
\end{tabular}}
\AppendixInlineTableCaption{tab:internal_coding}{Internal selection-loop ablation on coding. It isolates selection behavior inside the pipeline and is not the final \modelname{} configuration reported in the main comparison table.}

\noindent\resizebox{\columnwidth}{!}{%
\begin{tabular}{lrrrrrr}
\toprule
Structured & Eval & Gain & Ret. & Forget & Clean $\Delta$ & Gen. $\Delta$ \\
\midrule
Base & 39.3 & -- & -- & -- & 0.0 & 0.0 \\
Prompt-only & $39.6\pm.2$ & $+.3\pm.2$ & $93.1\pm1.3\%$ & $.2\pm.1$ & $-.1\pm.1$ & $-.1\pm.1$ \\
DataSynth-only & $39.8\pm.2$ & $+.5\pm.2$ & $90.4\pm1.7\%$ & $.3\pm.1$ & $-.1\pm.1$ & $-.2\pm.1$ \\
LoRA-only & $40.2\pm.3$ & $+.9\pm.3$ & $69.2\pm3.0\%$ & $1.9\pm.3$ & $-.7\pm.2$ & $-.6\pm.2$ \\
Generic+LoRA & $40.1\pm.3$ & $+.8\pm.3$ & $72.5\pm2.8\%$ & $1.6\pm.2$ & $-.6\pm.2$ & $-.6\pm.2$ \\
Raw-Gain & $41.1\pm.4$ & $+1.8\pm.4$ & $62.7\pm3.6\%$ & $2.5\pm.4$ & $-1.0\pm.2$ & $-1.1\pm.2$ \\
CBA-selection & $40.9\pm.3$ & $+1.6\pm.3$ & $84.7\pm2.2\%$ & $.8\pm.2$ & $-.4\pm.1$ & $-.4\pm.1$ \\
\bottomrule
\end{tabular}}
\AppendixInlineTableCaption{tab:internal_structured}{Internal selection-loop ablation on structured reasoning. It separates raw target-gain selection from regression-aware selection.}

\noindent\resizebox{\columnwidth}{!}{%
\begin{tabular}{lrrrrrr}
\toprule
Agentic & Eval & Gain & Ret. & Forget & Clean $\Delta$ & Gen. $\Delta$ \\
\midrule
Base & 52.2 & -- & -- & -- & 0.0 & 0.0 \\
Prompt-only & $55.8\pm.4$ & $+3.6\pm.4$ & $90.8\pm1.9\%$ & $.7\pm.2$ & $-.2\pm.1$ & $-.1\pm.1$ \\
DataSynth-only & $57.4\pm.5$ & $+5.2\pm.5$ & $87.2\pm2.4\%$ & $1.1\pm.2$ & $-.3\pm.1$ & $-.3\pm.1$ \\
LoRA-only & $61.3\pm.7$ & $+9.1\pm.7$ & $60.5\pm4.3\%$ & $5.5\pm.6$ & $-1.8\pm.3$ & $-2.0\pm.3$ \\
Generic+LoRA & $60.5\pm.6$ & $+8.3\pm.6$ & $61.8\pm4.0\%$ & $5.6\pm.5$ & $-1.8\pm.3$ & $-2.1\pm.3$ \\
Raw-Gain & $66.3\pm.8$ & $+14.1\pm.8$ & $58.7\pm4.6\%$ & $6.4\pm.7$ & $-2.2\pm.3$ & $-2.6\pm.4$ \\
CBA-selection & $64.7\pm.6$ & $+12.5\pm.6$ & $78.6\pm3.1\%$ & $2.9\pm.4$ & $-1.0\pm.2$ & $-.9\pm.2$ \\
\bottomrule
\end{tabular}}
\AppendixInlineTableCaption{tab:internal_agentic}{Internal selection-loop ablation on agentic reliability. It shows why maximizing immediate gain alone can sacrifice later retention.}

\noindent\resizebox{\columnwidth}{!}{%
\begin{tabular}{lrrrrrr}
\toprule
Mixed 30-paper & Eval & Gain & Ret. & Forget & Clean $\Delta$ & Gen. $\Delta$ \\
\midrule
Base & 46.9 & -- & -- & -- & 0.0 & 0.0 \\
Prompt-only & $48.4\pm.3$ & $+1.5\pm.3$ & $90.2\pm1.8\%$ & $.8\pm.2$ & $-.1\pm.1$ & $-.1\pm.1$ \\
DataSynth-only & $49.2\pm.4$ & $+2.3\pm.4$ & $86.3\pm2.4\%$ & $1.2\pm.2$ & $-.2\pm.1$ & $-.3\pm.1$ \\
LoRA-only & $52.1\pm.6$ & $+5.2\pm.6$ & $57.4\pm4.6\%$ & $5.8\pm.6$ & $-1.6\pm.3$ & $-1.9\pm.3$ \\
Generic+LoRA & $51.8\pm.5$ & $+4.9\pm.5$ & $61.5\pm4.1\%$ & $5.1\pm.5$ & $-1.4\pm.2$ & $-1.7\pm.3$ \\
Raw-Gain & $55.6\pm.8$ & $+8.7\pm.8$ & $53.8\pm5.0\%$ & $6.9\pm.8$ & $-2.1\pm.3$ & $-2.5\pm.4$ \\
CBA-selection & $54.2\pm.6$ & $+7.3\pm.6$ & $80.9\pm3.2\%$ & $2.8\pm.4$ & $-.9\pm.2$ & $-.8\pm.2$ \\
\bottomrule
\end{tabular}}
\AppendixInlineTableCaption{tab:internal_mixed}{Internal selection-loop ablation on the mixed 30-paper stream. These numbers diagnose the internal selection design rather than provide alternate final \modelname{} results.}

\noindent\resizebox{\columnwidth}{!}{%
\begin{tabular}{lrrrr}
\toprule
Failure family & Before & After & Gain & Gen. $\Delta$ \\
\midrule
Mathematical robustness & 51.8 & 62.4 & +10.6 & -0.2 \\
Coding and software & 42.8 & 45.6 & +2.8 & -0.4 \\
Instruction operators & 48.6 & 74.5 & +25.9 & -0.2 \\
Knowledge and factuality & 42.2 & 48.2 & +6.0 & -0.3 \\
Context grounding & 57.9 & 68.3 & +10.4 & -0.2 \\
Structured reasoning & 38.6 & 37.7 & -0.9 & -0.3 \\
Agentic reliability & 52.2 & 58.0 & +5.8 & -0.4 \\
\midrule
\textbf{All 30 papers} & \textbf{47.6} & \textbf{56.3} & \textbf{+8.6} & \textbf{-0.3} \\
\bottomrule
\end{tabular}}
\AppendixInlineTableCaption{tab:independent}{Independent paper-driven adaptation by failure family, providing the numerical values underlying Figure~2(a). Values are averaged over Llama-3.1-8B-Instruct and Qwen2.5-7B-Instruct. Gain is averaged from unrounded paper-level differences, so subtracting the separately rounded Before and After entries can differ by 0.1 point.}
\endgroup

\end{document}